\documentclass[a4paper]{article}
\usepackage[round]{natbib}
\usepackage{times}
\usepackage{latexsym}
\usepackage{url}
\usepackage{graphicx}
\usepackage{algorithm}
\usepackage[noend]{algorithmic}
\usepackage{covington}
\usepackage{multirow}
\usepackage{multicol}

\title{A PDTB-Styled End-to-End Discourse Parser}

\author{
  Ziheng Lin, Hwee Tou Ng, and Min-Yen Kan \\
  Department of Computer Science\\
  National University of Singapore\\
  13 Computing Drive\\
  Singapore 117417\\
  {\tt \{linzihen,nght,kanmy\}@comp.nus.edu.sg} 
}

\date{}

\begin{document}
\maketitle
\begin{abstract}
We have developed a full discourse parser in the Penn Discourse Treebank (PDTB) style. Our trained parser first identifies all discourse and non-discourse relations, locates and labels their arguments, and then classifies their relation types. When appropriate, the attribution spans to these relations are also determined. We present a comprehensive evaluation from both component-wise and error-cascading perspectives.
\end{abstract}

\section{Introduction}

A piece of text is often not to be understood individually, but understood by linking it with other text units from its context. These units can be surrounding clauses, sentences, or even paragraphs. A text becomes semantically well-structured and understandable when its text units are linked interstructurally from the bottom up.

Even when a text is well-structured, finding the discursive relationships that hold a text together automatically is difficult. In natural language processing (NLP), the process of understanding the internal structure of a text has been called {\em discourse analysis}, while the process of realizing the semantic relations in between text units has been called {\em discourse parsing}. Over the last couple of decades, researchers have proposed a number of discourse frameworks from different perspectives for the purpose of discourse analysis and parsing \citep{Mann-Thompson-88,Hobbs-90,Lascarides-Asher-93,Knott-Sanders-98,Webber-04}. However, designing and constructing such a discourse analyzer or parser has been a difficult task, partially attributable to the lack of any large annotated data set.

The Penn Discourse Treebank (PDTB) \citep{Prasad-et-al-08} is a recently released, discourse-level annotation on top of the Penn Treebank (PTB), which aims to fill this need. Providing a common platform for discourse researchers, it is the first annotation that follows the lexically grounded, predicate-argument approach, as proposed in Webber's framework \citeyearpar{Webber-04}.
In our work, we have (a) designed a parsing algorithm that performs discourse parsing in the PDTB representation, and (b) implemented an end-to-end system that reduces this algorithm to practice in a fully data driven approach. This system includes components that are novel as well as improved components from previous work.
To the best of our knowledge, this is the first parser that performs end-to-end discourse parsing in the PDTB style. 
The demo and source code of the parser have been released online\footnote{\url{http://wing.comp.nus.edu.sg/~linzihen/parser/}}.

\section{Related Work}


Mann and Thompson \citeyearpar{Mann-Thompson-88} proposed  Rhetorical Structure Theory (RST) which takes a nucleus-satellite view on rhetorical relations.
Marcu~\citeyearpar{Marcu-97} formalized an algorithm to parse an unrestricted text into its discourse tree using the RST framework. 
He made use of cue phrases to split a sentence into {\em elementary discourse units} ({\em edus}), designed algorithms that are able to recognize discourse relations with or without the signals of cue phrases, and proposed four algorithms for determining the valid discourse tree given the relations of adjacent {\em edus}.

Continuing this vein, Soricut and Marcu~\citeyearpar{Soricut-Marcu-03} introduced probabilistic models to segment a sentence into {\em edus}, and to derive their corresponding sentence-level discourse structure, using lexical and syntactic features. They experimented with their models using the RST Discourse Treebank (RST-DT) corpus \citep{Carlson-et-al-01}.


Recently, duVerle and Prendinger~\citeyearpar{duVerle-Prendinger-09} made use of a support vector machines (SVM) approach, using a rich set of shallow lexical, syntactic, and structural features, to train two separate classifiers on identifying the rhetorical structures and labeling the rhetorical roles drawn from the RST-DT.

With the advent of the larger PDTB, some recent work has attempted to recognize discourse relations and arguments in this newer corpus. 
Using syntactic features extracted from the parse trees, Pitler and Nenkova~\citeyearpar{Pitler-Nenkova-09} introduced a model that is able to disambiguate the discourse usage of connectives and recognize Explicit relations. Wellner and Pustejovsky~\citeyearpar{Wellner-Pustejovsky-07}, Elwell and Baldridge~\citeyearpar{Elwell-Baldridge-08}, and Wellner~\citeyearpar{Wellner-09} proposed machine learning approaches to identify the {\em head words} of the two arguments for discourse connectives. Although their method is capable of locating the positions of the arguments, it is not able to label the extent of these arguments. Machine learning approaches are used to identify Implicit relations ({\it i.e.}, discourse relations that are not signaled by discourse connectives such as {\em because}) in Pitler et al.~\citeyearpar{Pitler-et-al-09} and our previous work~\citep{Lin-et-al-09}. All of these research efforts in the PDTB can be viewed as isolated components of a full parser. Our work differs from these prior efforts in that we design a parsing algorithm that connects all sub-tasks into a single pipeline, and we implement this pipeline into an end-to-end parser in the PDTB style.

Component-wise, we introduce two novel approaches to accurately locate and label arguments, and to label attribution spans. We also significantly improve on the current state-of-the-art connective classifier with newly introduced features. 

\section{The Penn Discourse Treebank}

The Penn Discourse Treebank (PDTB) adopts a binary predicate-argument view on discourse relations, where the connective acts as a predicate that takes two text spans as its arguments. The span to which the connective is syntactically attached is called Arg2, while the other is called Arg1. The PDTB provides annotation for each discourse connective and its two arguments. Example~\ref{ex:explicit} shows one Explicit relation where the connective is
\underline{underlined}, Arg1 is {\it italicized} and Arg2 is {\bf bolded}. The number at the end (0214) shows which Wall Street Journal (WSJ) article this relation is from.

\begin{example}
\label{ex:explicit}
\underline{When} {\bf he sent letters offering 1,250 retired major leaguers the chance of another season}, {\it 730 responded}. (0214)
\end{example}

The PDTB also examined sentence pairs within paragraphs for discourse relations other than Explicit. Example~\ref{ex:implicit} shows an Implicit relation where the annotator inferred an implicit connective {\em accordingly}. Some relations are {\em alternatively lexicalized} by non-connective expressions. Example~\ref{ex:altlex} is such an AltLex relation with the non-connective expression {\em That compared with}. If no Implicit or AltLex relation exists between a sentence pair, annotators then checked whether an entity transition (EntRel) holds, otherwise no relation (NoRel) was concluded.

\begin{example}
\label{ex:implicit}
{\it ``I believe in the law of averages,''} \fbox{declared San Francisco batting coach} \fbox{Dusty Baker after game two.} \underline{Implicit} = {\sc accordingly} {\bf ``I'd rather see a so-so hitter who's hot come up for the other side than a good hitter who's cold.''} (2202) 
\end{example}


\begin{example}
\label{ex:altlex}
{\it For the nine months ended July 29, SFE Technologies reported a net loss of \$889,000 on sales of \$23.4 million.} \underline{AltLex} [{\bf That compared with}] {\bf an operating loss of \$1.9 million on sales of \$27.4 million in the year-earlier period.} (0229)
\end{example}

The PDTB also provides a three-level hierarchy of relation types. In this work, we follow our previous work~\citep{Lin-et-al-09} and focus on the Level 2 types. For each discourse relation ({\it i.e.}, Explicit, Implicit, or AltLex) the PDTB also provides annotation for the attribution ({\it i.e.}, the agent that expresses the argument) for Arg1, Arg2, and the relation as a whole. For example, the text span in the box in Example~\ref{ex:implicit} -- {\em declared San Francisco batting coach Dusty Baker after game two} -- is the attribution span for Arg1.

\section{System Overview}

We designed our parsing algorithm to mimic the annotation procedure performed by the PDTB annotators. Figure~\ref{fig:algo} shows the pseudocode.
The input to the parser is a free text $T$, whereas the output is the discourse structure of $T$ in the PDTB style. The algorithm consists of three steps which sequentially label Explicit relations, Non-Explicit relations, and attribution spans.

The first step is to identify discourse connectives, label their Arg1 and Arg2 spans, and recognize their Explicit relation types. First, the parser identifies all connective occurrences in $T$ (Line~2 in Figure~\ref{fig:algo}), and labels them as to whether they function as discourse connectives or not (Lines~3--4). If a connective occurrence $C$ is determined to be a discourse connective, its Arg1 and Arg2 spans are then identified, and the parser classifies the tuple ($C$, Arg1, Arg2) into one of the Explicit relation types (Lines~5--7). The second step then examines all adjacent sentence pairs within each paragraph. For each pair ($S_i$, $S_j$) that is not identified in any Explicit relation from Step 1, the parser then classifies the pair into EntRel, NoRel, or one of the Implicit/AltLex relation types (Lines~10--13).
Note that our parser follows the PDTB representation to ignore inter-paragraph relations, {\it i.e.}, it ignores the adjacent sentence pair in between two paragraphs. In Step 3, the parser first splits the text into clauses (Line~16), and for each clause $U$ that appears in any discourse relations ({\it i.e.}, Explicit, Implicit, and AltLex relations; EntRel and NoRel are non-discourse relations), it checks whether $U$ is an attribution span (Lines~17--19). In this step, the parser also follows the PDTB representation to only identify attribution spans appearing in discourse relations.

\begin{figure}[htb]
\begin{algorithmic}[1]
\REQUIRE a text $T$
\ENSURE a discourse structure of $T$
\STATE \COMMENT Step 1: label Explicit relations
\STATE Identify all connective occurrences in $T$
\FOR{each connective occurrence $C$}
    \STATE Label $C$ as disc-conn or non-disc-conn
    \IF{$C$ is disc-conn}
        \STATE Label Arg1 span and Arg2 span of $C$
        \STATE Label ($C$, Arg1, Arg2) as one of the Explicit relations
    \ENDIF
\ENDFOR
\STATE
\STATE \COMMENT Step 2: label Implicit, AltLex, EntRel, and NoRel relations
\FOR{each paragraph $P$ in $T$}
    \FOR{each adjacent sentence pair ($S_i$, $S_j$) in $P$}
        \IF{($S_i$, $S_j$) is not labeled as an Explicit relation in Step 1}
            \STATE Label ($S_i$, $S_j$) as EntRel, NoRel, or one of the Implicit/AltLex relations
        \ENDIF
    \ENDFOR
\ENDFOR
\STATE
\STATE \COMMENT Step 3: label attribution spans
\STATE Split $T$ into clauses
\FOR{each clause $U$}
    \IF{$U$ is in some Explicit/Implicit/AltLex relation from Step 1 or 2}
        \STATE Label $U$ as attr-span or non-attr-span
    \ENDIF
\ENDFOR
\end{algorithmic}
\caption{Pseudocode for the discourse parsing algorithm.}
\label{fig:algo}
\end{figure}

\begin{figure}[htb]
\centering 
\includegraphics[width=0.6\textwidth]{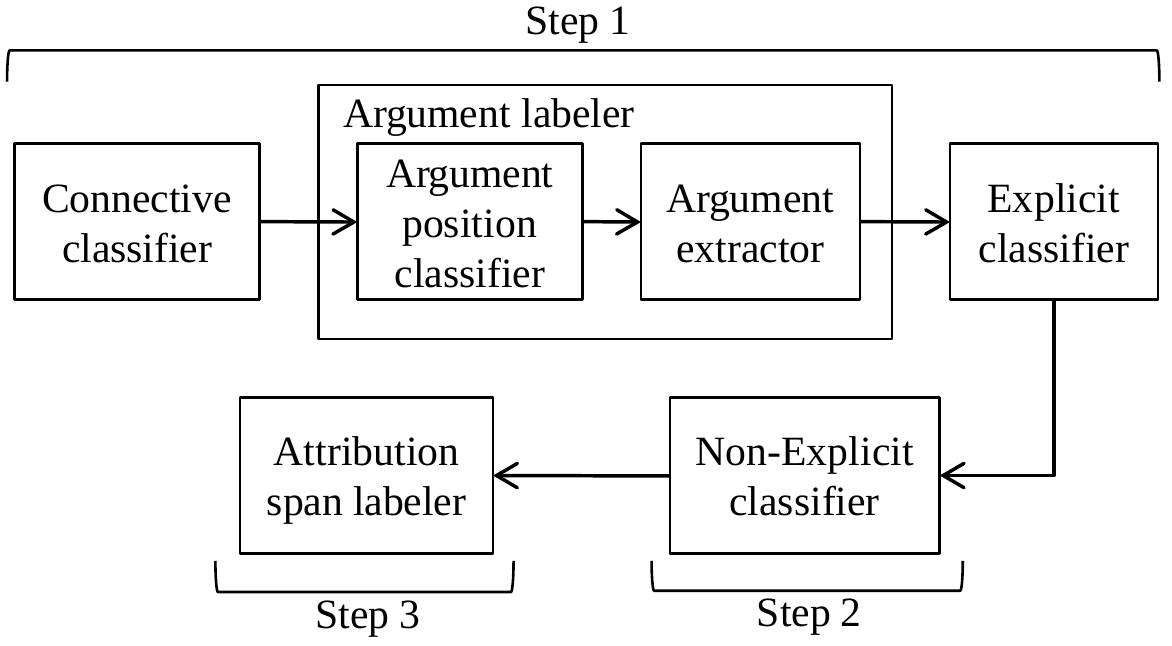}
\caption{System pipeline for the discourse parser.}
\label{fig:pipeline} 
\end{figure}

The pipeline of the parser is shown in Figure~\ref{fig:pipeline}, which consists of the connective classifier, argument labeler, explicit classifier, non-explicit classifier, and attribution span labeler. The first three components correspond to Step 1 in Figure~\ref{fig:algo}, while the last two correspond to Steps 2 and 3, respectively. There are two sub-components in the argument labeler: an argument position classifier and an argument extractor. A detailed description of these components follows in the next section.



\section{Components}

\subsection{Connective Classifier}

There are 100 types of discourse connectives defined in the PDTB. Given a connective occurrence such as {\em and}, the parser needs to decide whether it is functioning as a discourse connective. Pitler and Nenkova~\citeyearpar{Pitler-Nenkova-09} showed that syntactic features extracted from constituent parse trees are very useful in disambiguating discourse connectives. Beside the connective itself as a feature, they applied other syntactic features: the highest node in the tree that covers only the connective words (which they termed {\em self category}), the parent, left and right siblings of the self category, and two binary features that check whether the right sibling contains a VP and/or a trace. The best feature set they demonstrated also included pairwise interaction features between the connective and each syntactic feature, and the interaction features between pairs of syntactic features. 

In addition to the above, we observed that a connective's context and part-of-speech (POS) give a very strong indication of its discourse usage. 
For example, the connective {\em after} is usually functioning as a discourse connective when it is found followed by a present participle, as in ``after rising 3.9\%''. Based on this observation, we propose a set of lexico-syntactic features for a connective $C$ with its previous word $prev$ and next word $next$: $C$ POS, $prev$ + $C$, $prev$ POS, $prev$ POS + $C$ POS, $C$ + $next$, $next$ POS, and $C$ POS + $next$ POS. We also include as features the path from $C$ to the root, and the compressed path where adjacent identical tags are combined ({\it e.g.}, -VP-VP- is combined into -VP-).

\subsection{Argument Labeler}

The parser now labels the Arg1 and Arg2 spans of every discourse connective, in two steps: (1) identifying the locations of Arg1 and Arg2, and (2) labeling their extent. We note that Arg2 is the argument with which the connective is syntactically associated, and thus is fixed.
The remaining problem is in identifying the location of Arg1. We implement this as a classification task to recognize the {\em relative position} of Arg1, with respect to the connective. According to the different relative positions of Arg1, the argument extractor then attempts to extract the Arg1 and Arg2 spans. Figure~\ref{fig:arglab} gives the pseudocode for the argument labeler, which is further discussed in the following.

\begin{figure}[htb]
\begin{algorithmic}[1]
\REQUIRE a discourse connective $C$ and the text $T$
\ENSURE Arg1 and Arg2 spans of $C$ 
\STATE \COMMENT Argument position classifier
\STATE Classify the relative position of Arg1 as SS or PS
\STATE
\STATE \COMMENT Argument extractor
\IF{the relative position of Arg1 is SS}
\STATE Identify the Arg1 and Arg2 subtree nodes within the sentence parse tree
\STATE Apply tree subtraction to extract the Arg1 and Arg2 spans
\ELSE  [the relative position of Arg1 is PS]
\STATE Label the sentence containing $C$ as Arg2
\STATE Identify and label the Arg1 sentence from all previous sentences of Arg2
\ENDIF
\end{algorithmic}
\caption{Pseudocode for the argument labeler.}
\label{fig:arglab}
\end{figure}

\subsubsection*{Argument Position Classifier}

Prasad et al.~\citeyearpar{Prasad-et-al-08} described the demographic breakdown of the positions of Arg1 in their study of the PDTB annotations.  They showed that Arg1 can be located within the same sentence as the connective (SS), in some previous sentence of the connective (PS), or in some sentence following the sentence containing the connective (FS). PS is further divided into: in the immediately previous sentence of the connective (IPS) and in some non-adjacent previous sentence of the connective (NAPS). The distribution from their paper shows that 60.9\% of the Explicit relations are SS, 39.1\% are PS, and 0\% are FS (only 8 instances in the whole PDTB corpus). 

Motivated by this observation, we design an argument position classifier to identify the relative position of Arg1 as SS or PS. We ignore FS since there are too few training instances. We notice that the connective string itself is a very good feature. For example, when the connective token is {\em And} ({\it i.e.}, {\em and} with its first letter capitalized), it is a continuation from the previous sentence and thus Arg1 is likely in PS; whereas when the connective token is lowercase {\em and}, Arg1 is likely the clause at the left hand side of {\em and} and thus it is in SS. Additionally, some connectives always take a particular position. For example, {\em when} always indicates an SS case, whereas {\em additionally} always indicates PS.

%

Besides the connective string, we use the following contextual features in the classifier for the connective $C$ with its first and second previous words $prev_1$ and $prev_2$: position of $C$ in the sentence (start, middle, or end), $C$ POS, $prev_1$, $prev_1$ POS, $prev_1$ + $C$, $prev_1$ POS + $C$ POS, $prev_2$, $prev_2$ POS, $prev_2$ + $C$, and $prev_2$ POS + $C$ POS.

After the relative position of Arg1 is identified, the result is propagated to the argument extractor, which extracts the Arg1 and Arg2 spans accordingly.

\subsubsection*{Argument Extractor}

When Arg1 is classified as in the same sentence (SS), this means that Arg1, Arg2 and the connective itself are in the same sentence. This can be further divided into three cases: Arg1 coming before Arg2, Arg1 coming after Arg2, and Arg2 embedded within Arg1. One possible approach is to split the sentence into clauses before deciding which clause is Arg1 or Arg2. The problem with this approach is that it is not able to recognize the third case, where Arg2 divides Arg1 into two parts.

\begin{figure}[htb]
\centering
\includegraphics[width=0.85\textwidth]{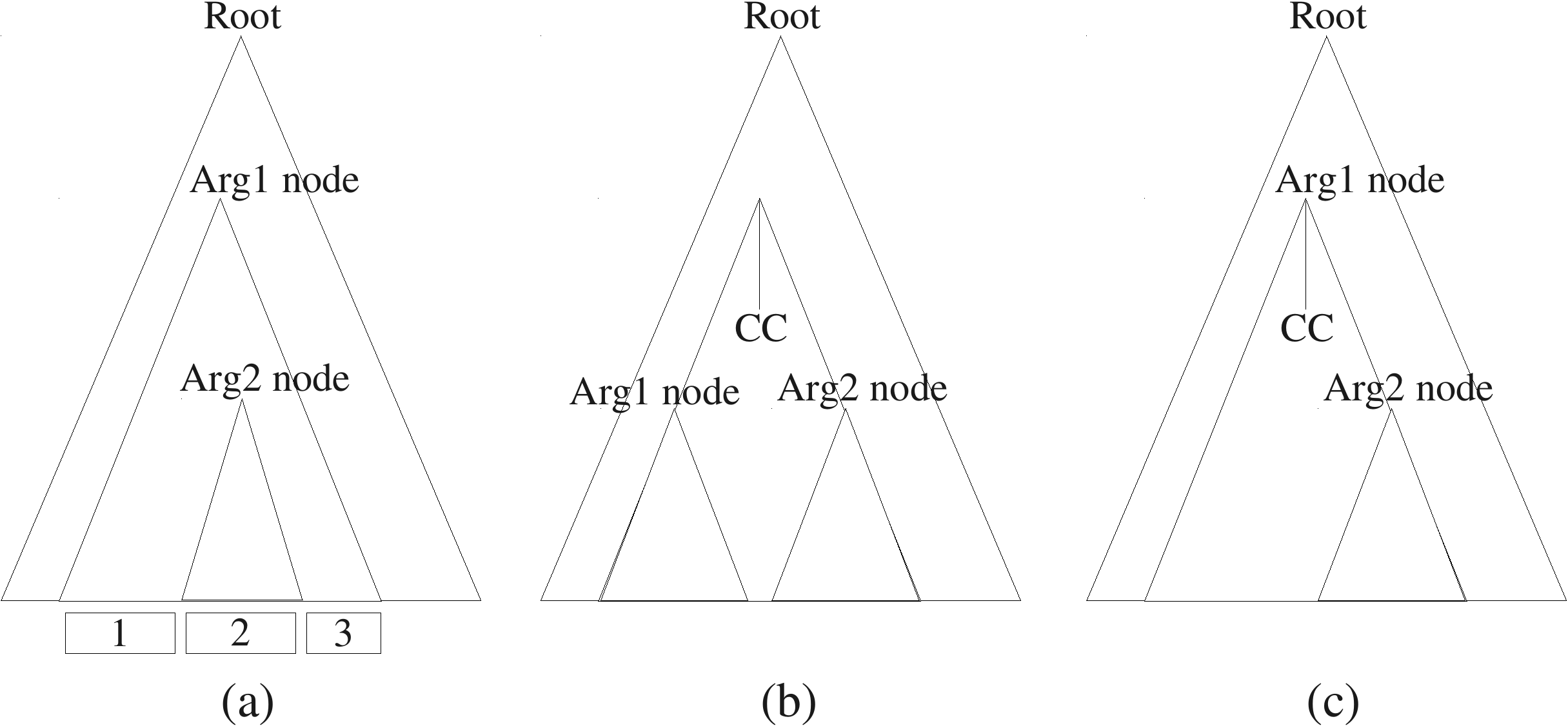}
\caption{Syntactic relations of Arg1 and Arg2 subtree nodes in the parse tree. Note that it is not possible for Arg1 node to be embedded in Arg2 node.}
\label{fig:tree-sub}
\end{figure}

Dinesh et al.~\citeyearpar{Dinesh-et-al-05} showed that Arg1 and Arg2 in the same sentence for subordinating connectives are always syntactically related as shown in Figure~\ref{fig:tree-sub}(a), where Arg1 and Arg2 nodes are the lowest nodes that cover the respective spans. They demonstrated that a rule-based algorithm is capable of extracting Arg1 and Arg2 in such cases for subordinating connectives. By using tree subtraction, the third case mentioned above can be easily recognized (span 2 in Figure~\ref{fig:tree-sub}(a) divides Arg1 into spans 1 and 3).

\begin{figure}[htb]
\centering
\includegraphics[width=0.35\textwidth]{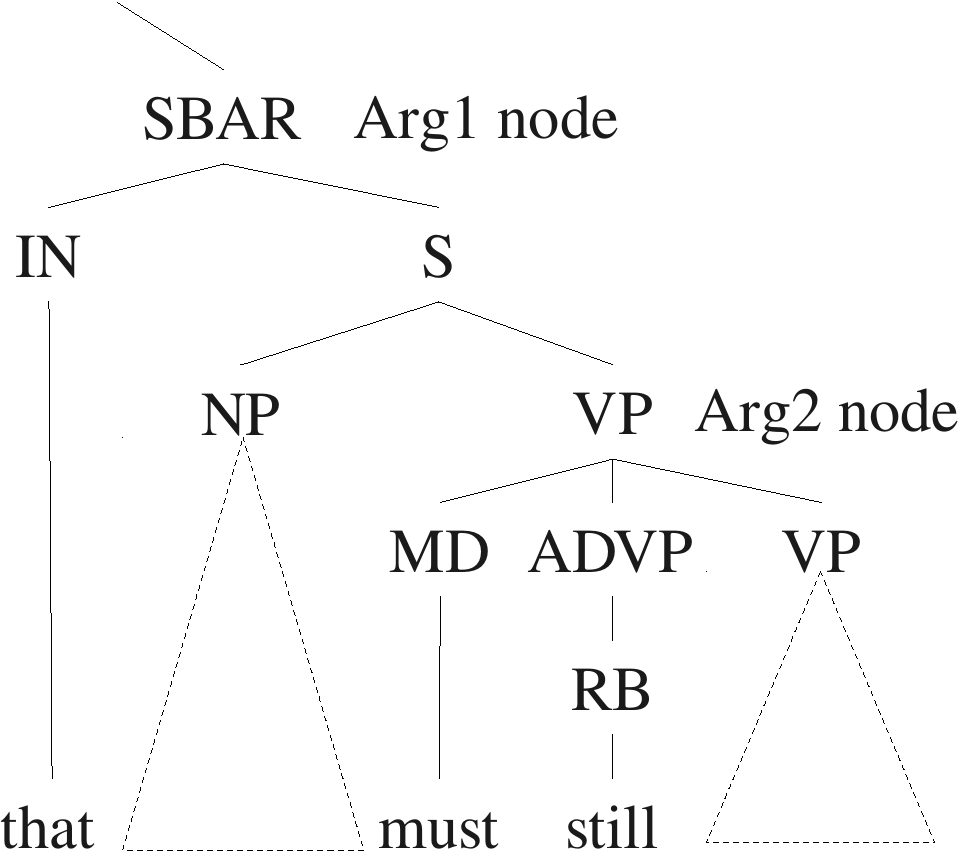}
\caption{Part of the parse tree for Example~\ref{ex:adv-ex} with Arg1 and Arg2 nodes labeled.}
\label{fig:adv-tree}
\end{figure}

However, dealing with only the subordinating connectives is not enough, because the percentages of coordinating connectives and discourse adverbials for SS cases occupy up to 37.50\% and 21.57\%, respectively, in the whole PDTB. We observe that coordinating connectives ({\em and}, {\em or}, {\em but}, etc.) usually constrain Arg1 and Arg2 to be syntactically related in one of two ways as shown in Figure~\ref{fig:tree-sub}(b)-(c), where CC is the connective POS. Discourse adverbials do not demonstrate such syntactic constraints as strongly as subordinating and coordinating connectives do, but their Arg1 and Arg2 are also syntactically bound to some extent. For example, Figure~\ref{fig:adv-tree} shows the syntactic relation of Arg1 and Arg2 nodes for the discourse adverbial {\em still} in Example~\ref{ex:adv-ex}.

\begin{example}
\label{ex:adv-ex}
Lilly, where the highest New York court expanded the market-share approach for the first time \fbox{to say} {\it that drug makers that could prove Mindy Hymowitz's mother didn't use their pill} {\bf must} \underline{still} {\bf pay their share of any damages}. (0130)
\end{example}

We design our argument node identifier to first identify the Arg1 and Arg2 subtree nodes within the sentence parse tree for all subordinating connectives, coordinating connectives and discourse adverbials, then apply tree subtraction to extract the Arg1 and Arg2 spans. The argument node identifier labels each internal node with three probabilities: functioning as Arg1-node, Arg2-node, and None. The internal node with the highest Arg1-node probability is chosen as the Arg1 node, and likewise for Arg2 node. The subtree under the Arg2 node is then subtracted from the Arg1 subtree to obtain the Arg1 spans, and the connective is subtracted from the Arg2 subtree to obtain the Arg2 span. Motivated by the syntactic properties observed, we propose the following features: the connective $C$, its syntactic category (subordinating, coordinating, or discourse adverbial), numbers of left and right siblings of $C$, path $P$ of $C$ to the node under consideration, the path $P$ and whether the size of $C$'s left sibling is greater than one, and the relative position of the node to $C$ (left, middle, or right). 
A maximum entropy classifier is used as it estimates class probabilities.

For the PS case where Arg1 is located in one of the previous sentences, the majority classifier labels the immediately previous sentence as Arg1, which already gives an $F_1$ of 76.90\% under gold standard setting in the whole PDTB. Since the focus of our work is not on identifying the Arg1 sentences for the PS case, we employ the majority classifier as our classifier.

\subsection{Explicit Classifier}
\label{subsec:exp}

After identifying a discourse connective and its two arguments, the next step is to decide what Explicit relation it conveys. 
Prasad et al.~\citeyearpar{Prasad-et-al-08} reported a human agreement of 94\% on Level~1 classes and 84\% on Level~2 types for Explicit relations over the whole PDTB corpus. The connective itself is a very good feature, as only a few connectives are ambiguous as pointed out in \citep{Miltsakaki-et-al-05}.
We train an explicit classifier using three types of features: the connective, the connective's POS, and the connective + its previous word. We follow our previous work~\citep{Lin-et-al-09} to train and test on the 16 Level 2 types. 

\subsection{Non-Explicit Classifier}

Besides annotating Explicit relations, the PDTB also provides annotation for Implicit relations, AltLex relations, entity transition (EntRel), and otherwise no relation (NoRel). We lump these together as Non-Explicit relations. The Non-Explicit relations are annotated for all adjacent sentence pairs within paragraphs. Furthermore, if there is already an Explicit relation between two adjacent sentences, their discourse relationship is already determined and are therefore exempt from further examination.

Similar to the explicit classifier, we adapt the Level~2 types for the Implicit and AltLex relations. As there are too few training instances for Condition, Pragmatic Condition, Pragmatic Contrast, Pragmatic Concession, and Exception relations, these five types are removed, resulting in 11 types.
Thus, our Non-Explicit classifier assigns candidate sentence pairs to one of 13 types (11 Level~2 types plus EntRel and NoRel).


We apply the four feature sets from our previous work~\citep{Lin-et-al-09}: contextual features (which check the existence of surrounding relations), constituent parse features, dependency parse features, and word-pair features. Besides these, we propose three new features to capture AltLex relations. AltLex relations are very similar to their counterpart Implicit relations, except that they are {\em alternatively lexicalized} by some non-connective expressions. We observe that such non-connective expressions are usually attached to the beginning of Arg2 ({\it e.g.}, such as ``That compared with'' in Arg2 of Example \ref{ex:altlex}). To distinguish AltLex relations, we use three features that take the first three words of Arg2 as their respective values. For the example above, the features will be {\em word$_1$=that}, {\em word$_2$=compared}, and {\em word$_3$=with}. 

\subsection{Attribution Span Labeler}

For each discourse relation ({\it i.e.}, Explicit, Implicit, or AltLex relation), the PDTB annotators labeled the attribution spans and annotated four dimensions for Arg1, Arg2, and the relation: their sources, types, scopal polarities, and determinacy. For the current parser, we develop a component to label the attribution spans, without labeling the four attribution dimensions and direction (Arg1, Arg2, or the relation) it is associated with. We follow the PDTB to only label attribution spans within discourse relations.

The attribution span labeler consists of two steps: splitting the text into clauses, and deciding which clauses are attribution spans. In the first step we employ a clause splitter that we have developed which uses syntactically motivated approach similar to \citep{Skadhauge-Hardt-05}. This clause splitter makes use of punctuation symbols and syntactic structures of SBAR complements.

The attribution span labeler then classifies each clause into attr-span or non-attr-span. We propose the following features extracted from the current, previous, and next clauses ($curr$, $prev$, and $next$): unigrams of $curr$, lowercased and lemmatized verbs in $curr$, the first and last terms of $curr$, the last term of $prev$, the first term of $next$, the last term of $prev$ + the first term of $curr$, the last term of $curr$ + the first term of $next$, the position of $curr$ in the sentence (start, middle, end, or whole sentence), and production rules extracted from $curr$.
Some clauses that belong to single attribution spans are incorrectly split into more than one clause by the clause splitter. For example, ``he said, adding'' is annotated as a single attribution span in the PDTB, but it is split into two clauses ``he said,'' and ``adding''. To correct such mistakes, after classification, adjacent attribution clauses within a sentence are combined to form a single attribution span.

\section{Evaluation}

In all of our experiments, we follow the recommendation from \citep{PDTB-07} to use Sec.~02--21 for training, Sec.~22 for development, and Sec.~23 for testing. All classifiers are trained with the OpenNLP maximum entropy package\footnote{\url{http://maxent.sourceforge.net/}}. 

For each component, the experiments are carried out when there is no error propagated from the previous components ({\it i.e.}, using gold standard annotation for the previous components), and when there is error propagation. As the PDTB was annotated on top of the PTB, we can either use the gold standard parse trees and sentence boundaries, or we can apply an automatic parser and sentence splitter. The experiments are carried out under three settings for each component: using gold standard parses and sentence boundaries (GS) without error propagation (EP), using GS with EP, and using both automatic parsing and sentence splitting (Auto) with EP. Thus GS without EP corresponds to a clean, per component evaluation, whereas the Auto with EP setting assesses end-to-end fully automated performance (as would be expected on new, unseen text input).


On the connective classifier, Pitler and Nenkova \citeyearpar{Pitler-Nenkova-09} (P\&N) reported an accuracy of 96.26\% and $F_1$ of 94.19\% with a 10-fold cross validation on Sec.~02--22. To compare with P\&N, we also run a 10-fold CV on Sec.~02--22 using their features and obtain replicated accuracy of 96.09\% and replicated $F_1$ of 93.57\%. Adding in our lexico-syntactic and path features, the performance is increased to 97.25\% accuracy and 95.36\% $F_1$, improvements of 0.99\% and 1.17\% over the reported results and 1.16\% and 1.79\% over the replicated results. A paired t-test shows that the improvements over the replicated results are significant with $p<0.001$\footnote{It is not possible to conduct paired t-test on the reported results for P\&N as we do not have the predictions.}.

In Table~\ref{tbl:conn}, we report results from the connective classifiers trained on Sec.~02--21 and tested on Sec.~23. The second and third columns show the accuracy and $F_1$ using the features of P\&N, whereas the last two columns show the results when we add in the lexico-syntactic and path features (+new). Introducing the new features significantly (all with $p<0.001$) increases the accuracy and $F_1$ by 2.04\% and 3.01\% under the GS setting, and 1.81\% and 2.62\% under the Auto setting.
This confirms the usefulness of integrating the contextual and syntactic information. As the connective classifier is the first component in the pipeline, its high performance is crucial to mitigate the effect of cascaded errors downstream.

When we look into the incorrectly labeled connectives, we find that the connective with the highest number of incorrect labels is {\em and} (8 false negatives and 4 false positives for the GS setting), which is not surprising, as {\em and} is always regarded as an ambiguous connective.

\begin{table}[htbp]
\centering
\begin{tabular}{|l|c|c|c|c|}
\hline 
    & \multicolumn{2}{|c|}{P\&N} & \multicolumn{2}{|c|}{+new} \\
\cline{2-5}
    & Acc.  & $F_1$ & Acc. & $F_1$ \\
\hline \hline
GS  & 95.30 & 92.75 & 97.34 & 95.76 \\
\hline
Auto& 94.21 & 91.00 & 96.02 & 93.62 \\
\hline
\end{tabular}
\caption{Results for the connective classifier.}
\label{tbl:conn}
\end{table}

We next perform evaluation on the argument position classifier, and report micro precision, recall and $F_1$, as well as the per class $F_1$. The GS + no EP setting gives a high $F_1$ of 97.94\%, which drops 3.59\% and another 2.26\% when error propagation and full automation are added in. The per class $F_1$ shows the performance degradation is mostly due to the SS class: the drops for SS are 5.36\% and 3.35\%, compared to 1.07\% and 0.68\% for PS. When we look into the contingency table for the GS + EP setting, we notice that out of the 36 false positives propagated from the connective classifier, 30 of them are classified as SS; for the Auto + EP setting there are 46 out of 52 classified as SS. This shows that the difference in the performance drops for SS and PS is largely due to the error propagation but not the classes themselves.

\begin{table}[htbp]
\centering
\begin{tabular}{|l|c|c|c|c|c|}
\hline 
            &       &       &     & \multicolumn{2}{|c|}{Per class $F_1$} \\
\cline{5-6}
            & Prec. & Recall& $F_1$ & SS & PS \\
\hline \hline
GS + no EP  & 97.94 & 97.94 & 97.94 & 98.26 & 97.49 \\
\hline
GS + EP     & 94.66 & 94.04 & 94.35 & 92.90 & 96.42 \\
\hline
Auto + EP   & 92.75 & 91.44 & 92.09 & 89.55 & 95.74 \\
\hline
\end{tabular}
\caption{Results for the argument position classifier.}
\label{tbl:argpos}
\end{table}

We next evaluate the performance of the argument extractor. Table~\ref{tbl:argnode} illustrates the results of identifying the Arg1 and Arg2 subtree nodes for the SS case for the three connective categories. The last column shows the relation level $F_1$ which requires both Arg1 and Arg2 nodes to be matched. We only show the results for the GS + no EP setting to save space. As expected, Arg1 and Arg2 nodes for subordinating connectives are the easiest ones to identify and give a high Arg2 $F_1$ of 97.93\% and a Rel $F_1$ of 86.98\%. We note that the Arg1 $F_1$ and Arg2 $F_1$ for coordinating connectives are the same, which is strange, as we expect Arg2 nodes to be handled more easily. The error analysis shows that Arg2 spans for coordinating connectives tend to include extra texts that cause the Arg2 nodes to move lower down in the parse tree. For example, ``... and Mr. Simpson said he resigned in 1988'' contains the extra span ``Mr. Simpson said'' which causes the Arg2 node moving two levels down the tree. As we discussed, discourse adverbials are difficult to identify as their Arg1 and Arg2 nodes are not strongly bound in the parse trees. However, as they do not occupy a large percentage in the test data, they do not lead to a large degradation as shown in the last row.

\begin{table}[htbp]
\centering
\begin{tabular}{|l|c|c|c|}
\hline 
            	& Arg1 $F_1$ 	& Arg2 $F_1$    & Rel $F_1$ 	\\
\hline \hline
Subordinating  	& 88.46     	& 97.93     	& 86.98 	\\
\hline
Coordinating    & 90.34     	& 90.34     	& 82.39 	\\
\hline
Discourse adverbial   & 46.88   & 62.50     	& 37.50 	\\
\hline
All  		& 86.63		& 93.41		& 82.60		\\
\hline
\end{tabular}
\caption{Results for identifying the Arg1 and Arg2 subtree nodes for the SS case under the GS + no EP setting for the three categories.}
\label{tbl:argnode}
\end{table}

Miltsakaki et al.~\citeyearpar{Miltsakaki-et-al-04} reported human agreements on both exact and partial matches to be 90.2\% and 94.5\%, respectively. They found that most of the disagreements for exact match come from partial overlaps which do not show significant semantic difference.
We follow such work and report both exact and partial matches. When checking exact match, we require two spans to match identically, excluding any leading and ending punctuation symbols. A partial match is credited if there is any overlap between the verbs and nouns of the two spans. The results for the overall performance for both SS and PS cases are shown in Table~\ref{tbl:argext}.
The GS + no EP setting gives a satisfactory $F_1$ of 86.24\% for partial matching on relation level. On the other hand, the results for exact matching are much lower than the human agreement. We observe that most misses are due to small portions of text being deleted from or added to the spans by the annotators to follow the {\em minimality principle} to include in the argument the minimal span of text that is sufficient for the interpretation of the relation, which poses difficulties for machines to follow.

\begin{table}[htbp]
\centering
\begin{tabular}{|l|l|c|c|c|}
\hline 
        &       & Arg1 $F_1$ & Arg2 $F_1$ & Rel $F_1$ \\
\hline \hline
\multirow{3}{*}{Partial} 
& GS + no EP    & 86.67     & 99.13     & 86.24 \\
\cline{2-5}
& GS + EP       & 83.62     & 94.98     & 83.52 \\
\cline{2-5}
& Auto + EP     & 81.72     & 92.64     & 80.96 \\
\hline \hline
\multirow{3}{*}{Exact} 
& GS + no EP    & 59.15     & 82.23     & 53.85 \\
\cline{2-5}
& GS + EP       & 57.64     & 79.80     & 52.29 \\
\cline{2-5}
& Auto + EP     & 47.68     & 70.27     & 40.37 \\
\hline
\end{tabular}
\caption{Overall results for the argument extractor.}
\label{tbl:argext}
\end{table}

Following the pipeline, we then evaluate the explicit classifier, with its performance shown in Table~\ref{tbl:exp}.
Recall that human agreement on Level 2 types is 84.00\% and a baseline classifier that uses only the connectives as features yields an $F_1$ of 86.00\% under the GS + no EP setting on Sec.~23. Adding our new features improves $F_1$ to 86.77\%.  With full automation and error propagation, we obtain an $F_1$ of 80.61\%. 
Pitler and Nenkova~\citeyearpar{Pitler-Nenkova-09} show that using the same syntactic features as their connective classifier is able to improve the explicit classifier on a 10-fold cross validation on Sec.~02-22. This actually performs worse than the baseline when trained on Sec.~02-21 and tested on Sec.~23.

\begin{table}[htbp]
\centering
\begin{tabular}{|l|c|c|c|}
\hline 
            & Precision & Recall    & $F_1$ \\
\hline \hline
GS + no EP  & 86.77     & 86.77     & 86.77 \\
\hline
GS + EP     & 83.19     & 82.65     & 82.92 \\
\hline
Auto + EP   & 81.19     & 80.04     & 80.61 \\
\hline
\end{tabular}
\caption{Results for the explicit classifier.}
\label{tbl:exp}
\end{table}

For the non-explicit classifier, a majority class baseline that labels all instances as EntRel yields an $F_1$ in the low 20s, as shown in the last column of Table~\ref{tbl:nonexp}. A single component evaluation (GS + no EP) shows a micro $F_1$ of 39.63\%. 
Although the $F_1$ scores for the GS + EP and Auto + EP settings are unsatisfactory, they still significantly outperform the majority class baseline by about 6\%. This performance is in line with the difficulties of classifying Implicit relations discussed in detail in our previous work~\citep{Lin-et-al-09}.

\begin{table}[htbp]
\centering
\begin{tabular}{|l|c|c|c|c|}
\hline 
            & Precision & Recall    & $F_1$     & Baseline $F_1$ \\
\hline \hline
GS + no EP  & 39.63     & 39.63     & 39.63     & 21.34 \\
\hline
GS + EP     & 26.21     & 27.63     & 26.90     & 20.30 \\
\hline
Auto + EP   & 24.54     & 26.45     & 25.46     & 19.31 \\
\hline
\end{tabular}
\caption{Results for the non-explicit classifier.}
\label{tbl:nonexp}
\end{table}

The final component, the attribution span labeler, is evaluated under both partial and exact match, in accordance with the argument extractor. From Table~\ref{tbl:attr}, we see that the GS + no EP setting achieves $F_1$ scores of 79.68\% and 65.95\% for the partial and exact match, respectively. When error propagation is introduced, the degradation of $F_1$ is largely due to the drop in precision. This is not surprising as at this point, the test data contains a lot of false positives propagated from the previous components. This has effect on the precision calculation but not recall (the recall scores do not change). When full automation is further added, the degradation is largely due to the drop in recall. This is because the automatic parser introduces noise that causes errors in the clause splitting step.

\begin{table}[htbp]
\centering
\begin{tabular}{|l|l|c|c|c|}
\hline 
        &       & Precision & Recall    & $F_1$ \\
\hline \hline
\multirow{3}{*}{Partial} 
& GS + no EP    & 79.40     & 79.96     & 79.68 \\
\cline{2-5}
& GS + EP       & 65.93     & 79.96     & 72.27 \\
\cline{2-5}
& Auto + EP     & 64.40     & 51.68     & 57.34 \\
\hline \hline
\multirow{3}{*}{Exact} 
& GS + no EP    & 65.72     & 66.19     & 65.95 \\
\cline{2-5}
& GS + EP       & 54.57     & 66.19     & 59.82 \\
\cline{2-5}
& Auto + EP     & 47.83     & 38.39     & 42.59 \\
\hline
\end{tabular}
\caption{Results for the attribution span labeler.}
\label{tbl:attr}
\end{table}

To evaluate the whole pipeline, we look at the Explicit and Non-Explicit relations that are correctly identified. We define a relation as correct if its relation type is classified correctly, and both its Arg1 and Arg2 are partially or exactly matched. Under partial matching, the GS + EP setting gives an overall system $F_1$ of 46.80\%, while under exact matching, it achieves an $F_1$ of 33.00\%. Auto + EP gives 38.18\% $F_1$ for partial match and 20.64\% $F_1$ for exact match. Expectedly, a large portion of the misses come from the Non-Explicit relations. The GS + EP results are close to the system $F_1$ of 44.3\% of an RST parser reported in~\citep{duVerle-Prendinger-09}. 


\section{Future Work}

In our explicit classifier, although the tuple ($C$, Arg1, Arg2) is passed into the classifier, the current approach does not make use of information from Arg1 and Arg2. One future work is to extract informative features from these two arguments for the explicit classifier. The current approach also does not deal with identifying Arg1 from all previous sentences for the PS case. We plan to design a PS identifier and integrate it into the current pipeline.

Wellner \citeyearpar{Wellner-09} pointed out that verbs from the attribution spans are useful features in identifying the argument head words. This suggests that we can feed back the results from the attribution span labeler into the argument labeler. In fact, we can feed back all results from the end of the pipeline into the start, to construct a joint learning model. 

\section{Conclusion}

In this work, we have designed a parsing algorithm that performs discourse parsing in the PDTB representation, and implemented it into an end-to-end system. This is the first end-to-end discourse parser that can parse any unrestricted text into its discourse structure in the PDTB style. We evaluated the system both component-wise as well as in an end-to-end fashion with cascaded errors. We reported overall system $F_1$ scores for partial matching of 46.80\% with gold standard parses and 38.18\% with full automation. We believe that such a discourse parser is very useful in downstream applications, such as text summarization and question answering (QA). For example, a text summarization system may utilize the contrast and restatement relations to recognize updates and redundency, whereas causal relations are very useful for a QA system to answer {\em why}-questions.

\bibliographystyle{plainnat}
\bibliography{tech}

\end{document}